\documentclass[10pt,twocolumn,letterpaper]{article}

\usepackage{iccv}
\usepackage{times}
\usepackage{epsfig}
\usepackage{graphicx}
\usepackage{amsmath}
\usepackage{amssymb}

\usepackage{pifont}
\usepackage{booktabs}
\usepackage{multirow}
\usepackage{xspace}
\usepackage[dvipsnames]{xcolor}

\usepackage{algorithm}
\usepackage{algpseudocode}
\usepackage[accsupp]{axessibility}

\definecolor{darkergreen}{RGB}{21, 152, 56}
\definecolor{red2}{RGB}{252, 54, 65}
\newcommand{\yesmark}{\textcolor{darkergreen}{\ding{52}}}
\newcommand{\nomark}{\textcolor{red2}{\ding{56}}}
\newcommand{\byesmark}{{\ding{52}}}
\newcommand{\bnomark}{{\ding{56}}}

\usepackage[pagebackref=true,breaklinks=true,letterpaper=true,colorlinks,bookmarks=false]{hyperref}

\usepackage[capitalize]{cleveref}

\newcommand{\oursfull}{\textbf{S}torage-\textbf{e}fficient V\textbf{i}sion \textbf{T}raining (\textbf{\ours})\xspace}
\newcommand{\ours}{SeiT\xspace}

\iccvfinalcopy %

\def\iccvPaperID{1251} %

\ificcvfinal\fi

\begin{document}

\title{SeiT: Storage-Efficient Vision Training with Tokens Using 1\% of Pixel Storage}

\author{Song Park$^*$ \quad Sanghyuk Chun$^*$ \quad Byeongho Heo \quad Wonjae Kim \quad Sangdoo Yun\\
{\small $^*$ Equal contribution}\\
{\small \,}
\\
{NAVER AI Lab}\\
}

\maketitle
\ificcvfinal\thispagestyle{empty}\fi

\begin{abstract}
We need billion-scale images to achieve more generalizable and ground-breaking vision models, as well as massive dataset storage to ship the images (\eg, the LAION-5B dataset needs 240TB storage space). However, it has become challenging to deal with unlimited dataset storage with limited storage infrastructure. A number of storage-efficient training methods have been proposed to tackle the problem, but they are rarely scalable or suffer from severe damage to performance. In this paper, we propose a storage-efficient training strategy for vision classifiers for large-scale datasets (e.g., ImageNet) that only uses 1024 tokens per instance without using the raw level pixels; our token storage only needs $<$1\% of the original JPEG-compressed raw pixels. We also propose token augmentations and a Stem-adaptor module to make our approach able to use the same architecture as pixel-based approaches with only minimal modifications on the stem layer and the carefully tuned optimization settings. Our experimental results on ImageNet-1k show that our method significantly outperforms other storage-efficient training methods with a large gap. We further show the effectiveness of our method in other practical scenarios, storage-efficient pre-training, and continual learning.
Code is available at \url{https://github.com/naver-ai/seit}
\end{abstract}

\section{Introduction}

We need billion-scale data points for more generalizable and ground-breaking vision models, \eg,  400M image-text pairs \cite{clip}, 1.8B image-text pairs \cite{jia2021scaling}, or 3.6B weakly-annotated images \cite{mahajan2018exploring, singh2022revisiting}. However, designing and operating a high-performance but fault-tolerant generic distributed dataset is a very expensive and challenging problem \cite{zaharia2012spark}. This problem has become more challenging for vision datasets compared to language datasets. For example, training GPT-2 with 8M documents only need 40GB of storage \cite{radford2019gpt2}, while the larger GPT-3 is trained with 410B tokens with 570GB of storage \cite{brown2020gpt3}. On the other hand, storing images requires significantly more storage space than storing language. For example, the ImageNet-21k dataset \cite{russakovsky2015imagenet} with 11M images requires a 1.4TB storage size, 2.5 times larger than GPT-3 storage despite containing fewer data points. Larger-scale datasets for large-scale pre-training require even more massive storage, \eg, 240TB for 5B images \cite{laion5b}. Consequently, storage remains a major bottleneck in scaling up vision models compared to language models.

\begin{figure}[t]
    \centering
    \includegraphics[width=\linewidth]{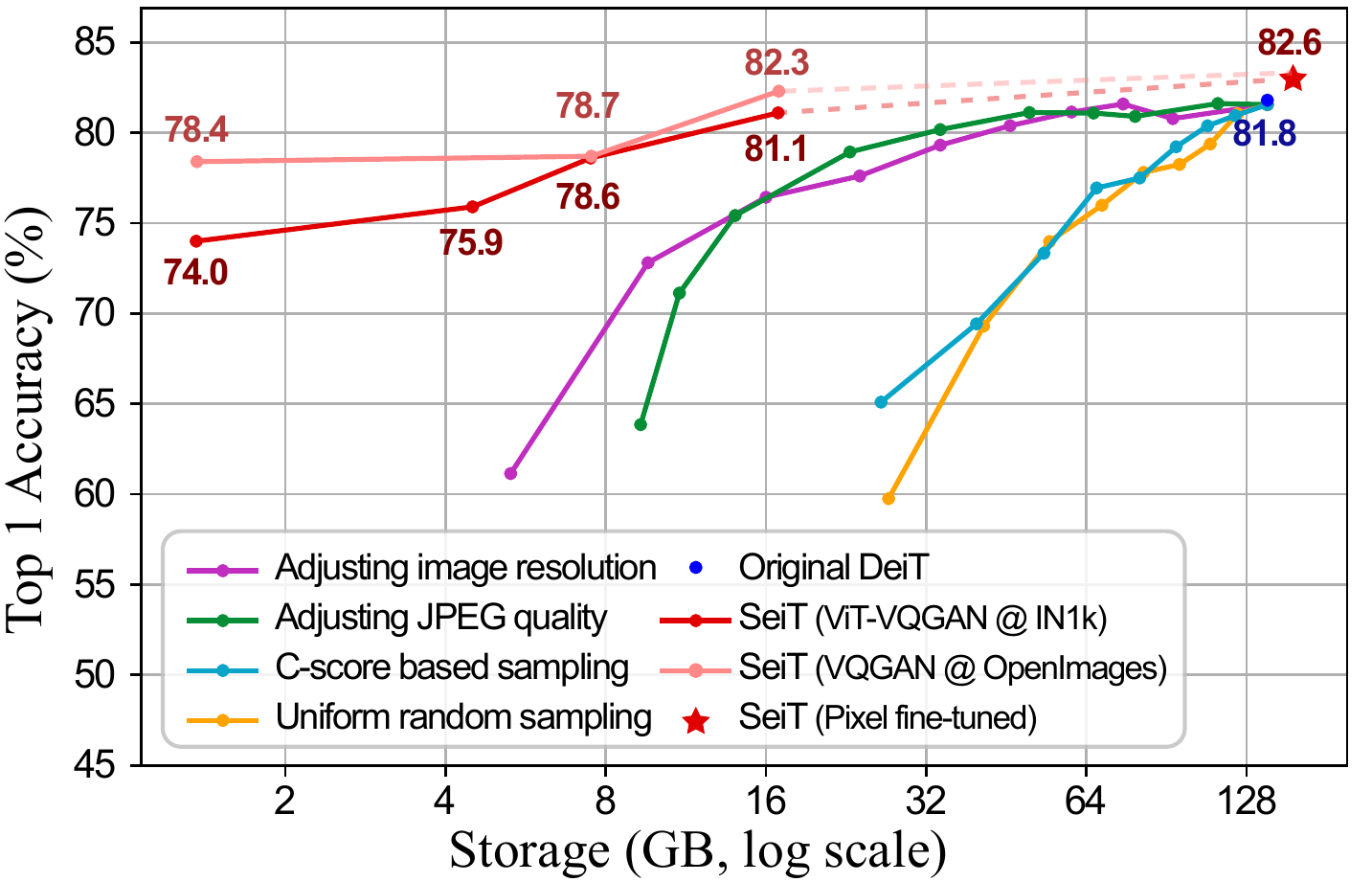}
    \caption{\small \textbf{Training data storage vs. ImageNet 1k Accuracy.}
    Comparisons on ImageNet-1k \cite{russakovsky2015imagenet} using ViT-B/16 backbone \cite{vit} are shown.
    Our \ours ({\color{red}red lines}) significantly outperforms other storage-efficient methods with the same storage size, achieving 74.0\% and 78.4\% top-1 acc with only 1.36GB utilizing tokenizers trained with ImageNet-1k and OpenImages, respectively. Note that the original pixel-based image storage requires 140GB of storage to achieve 81.8\% top-1 accuracy. Details are
    in \cref{tab:main_results_supp}.
    }
    \label{fig:teaser}
    \vspace{-.5em}
\end{figure}

Why do images require a large storage size than text? This is because while the nature of language is discrete, images are continuous in nature. 
Also, while the text quality is independent of document length, the image quality directly affects the storage size; better quality images require larger storage sizes. 
Although a lossy JPEG compression can reduce the storage size, as witnessed by Rombach \etal, still ``most bits of a digital image corresponds to imperceptible details'' \cite{rombach2022latentdiffusion}. Such imperceptible details (\eg, fine-grained details or high-frequency information of images) could be unnecessary for our desired vision classifiers. However, deep vision models are vulnerable to imperceptible high-frequency perturbations \cite{fgsm, pgd, autoattack} or unreasonably local areas \cite{geirhos2019stylized_imagenet, bahng2019rebias, scimeca2022wcst-ml}, implying that deep vision models attend too much to imperceptible details instead of the true property of objects. Therefore, we can expect that we can still achieve a high-performing vision model with the reduced image dataset by removing the imperceptible details.

There are two major directions to storage-efficient vision model training. The first direction aims to reduce the total number of data points by discarding less important samples \cite{coresets_for_efficient,2021mansheej_inportant_examples_diet,cscores} or synthesizing more ``condensed'' images than natural images \cite{data_condensation,data_condensation_augmentation}. However, this approach shows a significant performance drop compared to the full dataset (the blue and yellow lines in \cref{fig:teaser}) or cannot be applied to large-scale datasets due to their high complexity. Also, as the sampled or synthesized images are still normal images, these methods still suffer from an inefficient compression ratio to express imperceptible details. Furthermore, these methods need to compute the importance score or the sample-wise gradient of each sample by learning models with the full dataset. It makes these approaches not applicable to unseen datasets or newly upcoming data streams.

The other approach involves reducing the size of each image while keeping the total number of images. For example, by learning a more efficient compression method \cite{balle2016end, balle2018variational}. However, the neural compression methods have been mostly studied on extremely small-scale datasets (\eg, 24 images \cite{kodak} or 100 images \cite{asuni2014testimages}), and their generalizability to large-scale datasets is still an open problem. Moreover, the goal of neural compression is to compress an image and recover the original image as perfectly as possible, not to extract the most discriminant features for object recognition tasks. In response to these limitations, no neural compression method has been used to compress large-scale datasets like ImageNet \cite{russakovsky2015imagenet} to train deep vision models.

Due to the difficulty of the practical usage of neural compression, practitioners have attempted to reduce storage usage by controlling image quality. For example, the LAION dataset \cite{laion400m, laion5b} stores each image at 256 $\times$ 256 resolution, which takes up only 36\% of ImageNet images (469 $\times$ 387 resolution on average). Similarly, adjusting the JPEG compression quality can reduce the overall storage. As shown in \cref{fig:teaser} (green and purple lines), these approaches work well in practice compared to sampling-based methods. However, these methods have a limited compression ratio; if the compression ratio becomes less than 25\%, the performances drop significantly. By adjusting the image resolution with a 4\% compression ratio and JPEG quality with a 7\% compression ratio, we achieve 63.3\% and 67.8\% top-1 accuracies, respectively. In contrast, our approach achieves 74.0\% top-1 accuracy with only a 1\% compression ratio.

All shortcomings of the previous methods originate from the fact that too many imperceptible bits are assigned to store a digital image, which is misaligned with our target task. 
Our approach overcomes this limitation by storing images as tokens rather than pixels, using pre-trained vision tokenizers, such as VQGAN \cite{vqgan} or ViT-VQGAN tokenizer \cite{vitvqgan}. Introducing \oursfull, we convert each image to 32 $\times$ 32 tokens. The number of possible cases each token can have (the codebook) is 391, which takes only 1.15KB to store each token (assuming that the number of 391 cases can be expressed in 9 bits). It costs only less than 1.5GB for storing 140GB pixel-based storage of ImageNet. We train Vision Transformer (ViT) models on our tokenized images with minimum modifications. First, a 1024-length tokenized image is converted to a $32 \times 32 \times 32$ tensor by using pre-trained 32-dimensional codebook vectors from ViT-VQGAN. Next, we apply random resized crop (RRC) to the tensor to get a $32 \times 28 \times 28$ tensor. Then, to convert the tensor into a form that ViT can handle, we introduce \textit{Stem-Adapter} module that converts the RRC-ed tensor into a tensor of size $768 \times 14 \times 14$, the same as the first layer input of ViT after the stem layer. Because the image-based augmentations are not directly applicable to tokens, we propose simple token-specific augmentations, including \textit{Token-EDA} (inspired from easy data augmentation (EDA) \cite{wei2019eda} for language), \textit{Emb-Noise} and \textit{Token-CutMix} (inspired from CutMix \cite{yun2019cutmix}). In our experiment, we achieve 74.0\% top-1 accuracy with 1.36GB token storage, where the full image storage requires 140GB to achieve 81.8\% \cite{deit}.

\ours has several advantages over previous storage-efficient methods. First, as we use a frozen pre-trained tokenizer that only requires forward operations to extract tokens from images, we do not need an additional optimization for compressing a dataset, such as importance score-based sampling \cite{cscores}, image synthesis methods \cite{data_condensation, data_condensation_augmentation}, or neural compression \cite{balle2016end, balle2018variational}. Hence, \ours is easily applicable to newly upcoming data streams directly. Second, unlike previous works that use pre-trained feature extractors (\eg, HOG \cite{dalal2005hog} or Faster-RCNN \cite{ren2015fasterrcnn, anderson2018bottomup}), \ours can use the same architecture as pixel-based approaches with only minimal modifications on the stem layer, as well as the carefully tuned optimization settings, such as DeiT \cite{deit}. It becomes a huge advantage when using \ours as an efficient pre-training method; we can achieve 82.6\% top-1 accuracy by fine-tuning the token pre-trained model with images. Moreover, applying an input augmentation for feature extractor-based approaches is not straightforward, limiting their generalizability. Finally, \ours shows a significant compression ratio, with a 1\% compression ratio for ImageNet.

We show the effectiveness of \ours on three image classification scenarios: (1) storage-efficient ImageNet-1k benchmark (2) storage-efficient large-scale pre-training, and (3) continual learning. The overview of storage-efficient results is shown in \cref{fig:teaser}: \ours outperforms comparison methods with a significant gap with the same storage size, 74.0\% accuracy on ImageNet under 1\% of the original storage, where comparison methods need 40\% (uniform sampling, C-score sampling \cite{cscores}), 6\% (adjusting image resolution), and 8\% (adjusting JPEG quality) of the original storage to achieve the similar performance. We also demonstrate that \ours can be applied to large-scale pre-training for an image-based approach; we pre-train a ViT-B/16 model on the tokenized ImageNet-21k (occupying only 14.1GB) and fine-tune the ViT model on the full-pixel ImageNet-1k. By using slightly more storage (156GB vs. 140GB), our storage-efficient pre-training strategy shows 82.8\% top-1 accuracy, whereas the full-pixel ImageNet-1k training shows 81.8\%. Finally, we observe that our token-based approach significantly outperforms the image-based counterpart in the continual learning scenario \cite{rolnick2019experience_replay_er} by storing more data samples in the same size of the memory compared to full-pixel images.

\paragraph{Contributions.} (1) We compress an image to 1024 discrete tokens using a pre-trained visual tokenizer. By applying a simple lossless compression for the tokens, we achieve only 0.97\% storage size compared to images stored in pixels. (2) We propose Stem-Adapter module and augmentation methods for tokens such as Token-RRC, Token-CutMix, Emb-Noise, and Token-EDA in order to enable ViT training with minimal change to the protocol and hyperparameters of existing ViT training. (3) Our storage-efficient training pipeline named \oursfull shows great improvements on the low-storage regime. With only 1\% storage size, \ours achieves 74.0\% top-1 ImageNet 1k validation accuracy. (4) We additionally show that \ours can be applied to a storage-efficient pre-training strategy, and continual learning tasks.

\section{Related Works}

\paragraph{Importance sampling for efficient training.}
Sampling-based methods \cite{coresets_for_efficient,2021mansheej_inportant_examples_diet,coleman2019selection,cscores} aims to idendity a compact, yet representative subset of the training dataset that satisfies the original objectives for efficient model training. This is usually achieved through exploring the early training stage \cite{2021mansheej_inportant_examples_diet}, constructing a proxy model \cite{coleman2019selection}, or utilizing consistency score (C-score) \cite{cscores}. However, the empirical performance gap between sampling-based methods and the baseline approach of random selection is insignificant, particularly in large-scale datasets like ImageNet-1k (See \cref{fig:teaser}). We believe that preserving the diversity of data points in a dataset is crucial, and therefore we endeavor to maintain the number of data points instead of pruning them.

\paragraph{Dataset distillation.}
Dataset distillation \cite{wang2018dataset} aims to generate a compact dataset by transferring the knowledge of the training dataset into a smaller dataset. Recent works \cite{data_condensation, data_condensation_augmentation, lee2022dcc, sangermano2022sample, rosasco2022distilled} have shown that the synthesized data can be effective in efficient model training, especially in scenarios such as continual learning \cite{rolnick2019experience_replay_er}. However, due to their high complexity, they have not yet demonstrated successful cases in large-scale datasets such as ImageNet-1k. We recommend the survey paper \cite{lei2023dataset_distillation_survey} for curious readers.

\paragraph{Neural compression.}
Image compression algorithms have improved with the use of neural network training to minimize quality loss on lossy compression. 
The representative learned image compression methods are based on VAE \cite{balle2016end, balle2018variational}. The compressor encodes an image to discrete latent codes and the codes can be decoded into the image with small losses. Recent studies \cite{cheng2020learned,kim2022joint} have utilized the self-attention mechanism \cite{vaswani2017attention} with heavy CNN architectures to demonstrate superior compression power compared to conventional methods such as JPEG. However, the learned image compression targets high-quality images with complex and detailed contexts, which are distant from ImageNet samples. Thus, it is challenging to apply these methods to compress ImageNet for ViT training.

\paragraph{Learning with frozen pre-extracted features.}
Using extracted visual features for a model has been widely used in the computer vision field. It shows reasonable performances with a low computational cost compared to pixel-based visual encoders. For example, the Youtube-8M \cite{abu2016youtube} dataset consists of frame features extracted from Inception \cite{szegedy2015going} instead of raw pixels, allowing efficient video model training \cite{mao2018hierarchical,bhardwaj2019efficient} with frozen frame features. The pre-extracted features have also been widely used for tasks that need higher knowledge than pixel-level understandings. For example, frozen CNN features \cite{lu2016hierarchical} or bottom-up and top-down (BUTD) \cite{teney2018tips,anderson2018bottomup} features \cite{kim2018bilinear} have been a popular choice for vision-and-language models that aim to understand complex fused knowledge between two modalities, \eg, visual question answering \cite{antol2015vqa,goyal2017making}. These approaches show slightly worse performances than the end-to-end training from raw inputs without pre-extracted features \cite{kim2021vilt, clip}, but show high training efficiency in terms of computations.

However, these methods need feature-specific modules to handle frozen features and specialized optimization techniques rather than standard optimization methods of pixel-based methods. Furthermore, some fundamental augmentations, such as random resized crop (RRC), are not applicable to the frozen features, resulting in inferior generalizability. \ours has major advantages over these methods where it is the almost same training method for ViT (\eg, DeiT \cite{deit}), and yet it can significantly reduce the storage space.

\section{Token-based Storage-Efficient Training}

In this section, we propose \oursfull. \ours aims to learn a high-performing vision classifier at scale (\eg, ImageNet-1k \cite{russakovsky2015imagenet}) with a small storage size (\eg, under 1\% of the original size), a minimal change on the training strategy (\eg, highly optimized training strategy \cite{deit}), and the minimum sacrifice of accuracies. \ours consists of two parts (1) preparing the compressed token dataset and (2) training a model using the tokens.

\begin{figure}[t]
    \centering
    \includegraphics[width=0.9\linewidth]{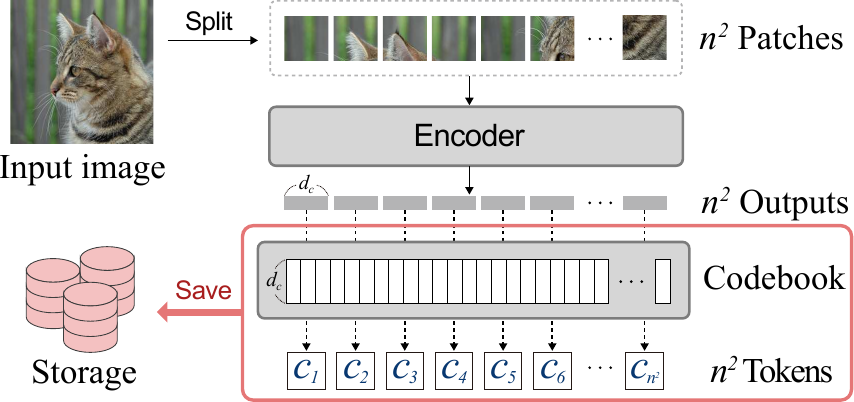}
    \caption{\small \textbf{Tokenization.} The input image is resized to 256 $\times$ 256 and then divided into non-overlapping $n^2$ patches. The patches are fed into the ViT-VQGAN encoder, which produces a sequence of $d_c$ dimensional vectors from the patches. 
    Finally, the tokens are generated by mapping each vector to the nearest code in a pre-trained codebook. We used 32 for both $n$ and $d_c$ in this paper.}
    \label{fig:tokenization}
\end{figure}

\begin{figure*}
    \centering
    \includegraphics[width=\linewidth]{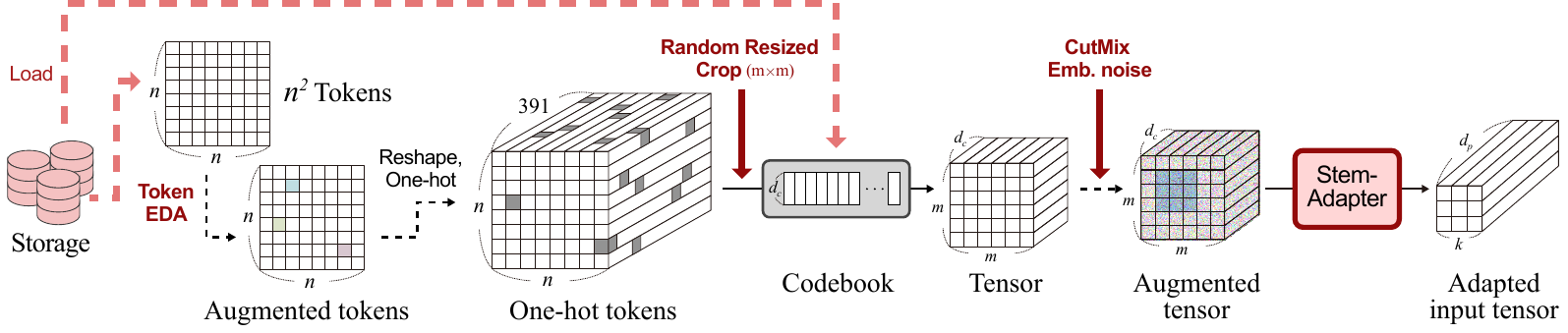}
    \caption{\small \textbf{The data processing pipeline for token data.} For each image, a $n \times n$-shaped token is loaded from the storage. We apply Token-EDA to augment the token and convert it into a one-hot form. Then, we randomly resize crop the one-hot token to $m \times m$ and process it using the pre-trained ViT-VQGAN codebook to transform it into a $d_c \times m \times m$ tensor. We further apply CutMix and Embedding-noise to this tensor and the augmented tensor is then fed into the Stem-Adapter module, which transforms it into a shape of $d_V \times k \times k$, making it suitable for use with ViT models. Our experimental values for the parameters are $n=32$, $m=28$, $k=14$, $d_c=32$, and $d_V=768$.}
    \label{fig:augmentation}
\end{figure*}

\subsection{Preparing the token dataset}
We extract tokens using the ImageNet-trained ViT-VQGAN tokenizer \cite{vitvqgan} because it shows the best reconstruction quality among the ImageNet-1k only trained tokenizers (See Appendix). In \cref{fig:teaser} and Appendix, our approach performs better if a stronger tokenizer trained with an extra dataset, \eg, the OpenImages-trained VQGAN tokenizer \cite{vqgan}, is used. In the main paper, however, we use the ViT-VQGAN tokenizer for a fair comparison with other storage-efficient methods in terms of the training dataset.

\cref{fig:tokenization} shows the overview of the dataset preparation pipeline.
We first resize the entire ImageNet dataset to 256 $\times$ 256. Then, each resized image is divided into non-overlapping 8 $\times$ 8 image patches. Finally, we encode each patch into a 32-dimensional vector and assign a code index by finding the nearest codeword from the pre-trained codebook. Here, we only use 391 codewords from the 8192 original codewords because we found that only 391 codewords are used for the ImageNet training dataset. As a result, each image is converted to 32 $\times$ 32 tokens where each token belongs to [0, $\ldots$, 390]. We also store the codebook of ViT-VQGAN (a 32 $\times$ 391 vector) to re-use the knowledge of the codebook for better performance.

\begin{table}[t]
\small
\centering
\begin{tabular}{llrr}
\toprule
Format & Encoding & \begin{tabular}[c]{@{}c@{}}Storage\\ size \end{tabular} & \begin{tabular}[c]{@{}c@{}} Avg. size\\ per image\end{tabular} \\
\midrule
Pixels & \texttt{uint8} (uncompressed) & 1471.2 GB & 1.14 MB  \\
Pixels & JPEG (baseline) & 140.0 GB & 109.3 kB  \\
Tokens & \texttt{uint16} (uncompressed) & 2.50 GB & 2.0 kB\\
Tokens & Ours (8 bits encoding) & 1.54 GB & 1.26 kB\\
Tokens & Ours + Huffman coding  & \textbf{1.36 GB} & \textbf{1.11 kB}\\
\midrule
Tokens & \textit{Theoretical optimum} & 1.32 GB & 1.08 kB\\
\bottomrule
\end{tabular}
\vspace{.5em}
\caption{\small {\bf Storage size of the ImageNet-1k training dataset for different formats and encodings.} \texttt{uint8} and \texttt{uint16} denote uncompressed version of each data format. \textit{Theoretical optimum} is estimated by assuming the token population is uniform.}
\label{tab:format_storage}
\end{table}

In theory, as our token indices belong to [0-390], the optimal bit length to store the tokens is $\log_2 391 = 8.61$\footnote{Following the empirical population of the tokens, the ``empirical'' optimal bit length is 8.54 by computing $H(p) = - \sum p_i \log p_i$. However, in the rest of the paper, we assume the population is uniform for simplicity.} by the source coding theorem \cite{cover1999information}. Therefore, the optimal storage size of an image will be 1.08 kB\footnote{We have 1.08 kB = bits per token (8.61) $\times$ token length (1024) / bits per Byte (8). If we follow the actual distribution, it becomes 1.07 kB.}. However, in practice, we cannot store tokens in 8.61 bits because the commonly used data types use Byte for the minimal unit, \eg, 1 Byte (\texttt{uint8}) or 2 Bytes (\texttt{uint16}). To compress the required bits per token to less than 2 Bytes, we propose a simple yet efficient encoding for the tokens. First, we assign each token index following the token popularity, \ie, the most frequent token is assigned to index 0, and the least frequent token is assigned to index 390. Then, we break up token indices larger than 255 into two elements as follows:
\begin{equation}
\label{eq:encoding}
    i = \begin{cases}
        [i] \quad\quad~\text{if}~ i < 255\\
        [255, i] ~\text{if}~ i \geq 255\\
    \end{cases}
\end{equation}
We store multiple tokens in a file to reduce the required storage as small as possible. However, because our encoding process makes the length of each token variable, the naive decoding process for our encoding will need $O(n)$ complexity where $n$ is the number of encoded tokens by \cref{eq:encoding}. We solve the problem by simply storing the start indices of each image. The index storage only requires 9.8 MB for the entire ImageNet training dataset, but it makes the decoding process becomes $O(1)$ and parallelizable. 
Pseudo-codes for the proposed encoding-decoding are in \cref{subsec:pseudocode_supp}.

Our simple encoding strategy reduces 40\% of the overall storage size compared to the naive \texttt{uint16} data type as shown in \cref{tab:format_storage}. Here, as the original baseline storage also employs a compression algorithm, such as JPEG (See the first and the second row of \cref{tab:format_storage}), we also apply a simple compression algorithm, Huffman coding \cite{huffman1952method}. After applying Huffman coding to our token storage, we achieve nearly optimal storage size per image (1.11 kB vs. 1.08 kB). We empirically observe that the entire decoding process, including Huffman decoding, is almost neglectable: while the full-pixel processing requires 0.84s per 100 images, our approach only needs 0.07s.
As a result, full-pixel and SeiT take 5m 40s and 5m 12s for 1 epoch training, respectively.
In the remaining part of this paper, we use the compressed version of our token dataset if there is no specification.

\subsection{Training classifiers with tokens}

Training a classifier with tokenized images is not trivial. For example, an input token has 32 $\times$ 32 dimensions, but a conventional image input has 3 $\times$ 224 $\times$ 224. Furthermore, strong image-level augmentations (\eg, RandAugment \cite{cubuk2020randaugment}, Gaussian blur \cite{touvron2022deit}) have become crucial in large-scale vision classifiers, however, these operations cannot be directly applied to the tokens. One possible direction is to decode tokens to pixel-level images during every forward computation. However, this would impose an additional computational load on the network. 
Instead, we propose simple yet effective token-level augmentations and a simple Stem-Adapter module to train a vision classifier directly on the tokens with minimal modification but small sacrifices.

\subsubsection{Token Augmentations}

\noindent \textbf{Token-EDA.}
We utilize the EDA \cite{wei2019eda}, designed for language models, to augment our token data. EDA originally involves four methods: Synonym Replacement (SR), Random Insertion (RI), Random Swap (RS), and Random Deletion (RD). However, we only adopt SR and RS because the others do not maintain the number of tokens, which is not compatible with the ViT training strategy. For SR, we define synonyms of a token as the five tokens that have the closest Euclidean distance in the ViT-VQGAN codebook space. Then, each token is randomly replaced with one of its synonyms with a certain probability $p_{s}$ during training. For RS, we randomly select two same-sized squares from a 32 $\times$ 32 token and swapped the tokens inside them with each other, with a probability $p_{r}$. We use 0.25 for $p_{s}$ and $p_{r}$ for \ours.

\noindent \textbf{Token-RRC and Token-CutMix.}
In addition to EDA, we apply Random Resized Crop (RRC) and CutMix \cite{yun2019cutmix} to tokens. For RRC, we adopt a standard ImageNet configuration with a scale (0.08, 1) and an aspect ratio (3/4, 4/3). To enable interpolation, we first convert the original 32 $\times$ 32 tokens to one-hot form. Then, apply the random cropping to these one-hot tokens, which are subsequently resized to 28 $\times$ 28 using bicubic interpolation. After RRC, the one-hot tokens are converted to a 32 $\times$ 28 $\times$ 28 tensor using the pre-trained codebook vectors from ViT-VQGAN, where 32 is the size of a pre-trained code vector. Note that tokens that are not in one-hot form due to interpolation are converted to mixed codebooks following their values. CutMix is then applied to these tensors, whereby a patch is randomly selected from one token and replaced with a patch from another token while maintaining the channel dimension.

\begin{table*}[]
\small
\centering
\begin{tabular}{ccrrrr}
\toprule

\multicolumn{1}{c}{Method}                & \begin{tabular}[c]{@{}c@{}} Reduction\\ factor\end{tabular} & \begin{tabular}[c]{@{}c@{}} Dataset\\ storage size\end{tabular} & \# of images & \begin{tabular}[c]{@{}c@{}} Avg. size\\ per image\end{tabular} & Top1 Acc. \\
\midrule
\multicolumn{1}{c}{Full-pixels}           & 100\% & 140.0 GB    & 1,281 k   & 109 kB         & 81.8      \\
\midrule
\multirow{3}{*}{Uniform random sampling}   & 70\% & 95.7 GB   & 897 k    & 107 kB       & 78.2      \\
                                           & 40\% & 54.6 GB   & 512 k   & 107 kB        & 74.0        \\
                                           & 20\% & 27.2 GB   & 256 k   & 107 kB        & 59.8      \\
                                           \midrule
\multirow{3}{*}{C-score~\cite{cscores} based sampling}    & 60\% & 80.6 GB   & 769 k   & 105 kB        & 77.5      \\
                                           & 40\% & 53.3 GB   & 512 k   & 104 kB        & 73.3      \\
                                           & 20\% & 26.3 GB   & 256 k   & 103 kB        & 65.1      \\
                                           \midrule
\multirow{3}{*}{Adjusting image reolution} & 30\% & 16.0 GB   & 1,281 k    & 13 kB        & 78.6      \\
                                           & 20\% & 9.6 GB    & 1,281 k    & 8 kB        & 75.2      \\
                                           & 10\% & 5.3 GB    & 1,281 k    & 4 kB        & 63.3      \\
                                           \midrule
Adjusting JPEG quality factor & 10   & 14.0 GB    & 1,281 k    & 11 kB        & 78.1      \\

(an integer scale between 1-100 representing & 5    & 11.0 GB    & 1,281 k    & 9 kB        & 74.6      \\
particular compression levels) & 1    & 9.3 GB    & 1,281 k    & 7 kB        & 67.8      \\
                                           \midrule
\multicolumn{1}{c}{\ours (ImageNet-1k-5M \cite{yun2021re}, the full dataset)}  & -    & 7.5 GB   & 5,830 k    & 1 kB            & 78.6      \\
\multicolumn{1}{c}{\ours (ImageNet-1k-5M, 60\% randomly sampled one)}     & 60\%    & 4.5 GB    & 3,498 k    & 1 kB         & 75.9       \\
\multicolumn{1}{c}{\ours (ImageNet-1k, the full dataset)}     & -    & 1.4 GB    & 1,281 k    & 1 kB         & 74.0        \\

\bottomrule
\end{tabular}
\vspace{.5em}
\caption{\small \textbf{Main results.} ImageNet-1k results using various data storage reduction methods are shown. We compare \ours against reduction factors that achieve comparable performance and storage size.
Note that the numbers for all reduction factors are included in \cref{subsec:full_exp_supp}.
}
\label{tab:main_results}
\end{table*}

\noindent \textbf{Adding channel-wise noise.}
We also developed Emb-Noise, a token augmentation method that mimics color-changing image augmentations, such as color jittering. Inspired by the fact that each channel in an image represents a specific color, we first generate noise of length 32 and add it to each channel of the converted tensor with 32 $\times$ 28 $\times$ 28 dims, and then apply full-size iid noise, i.e. noise size of 32 $\times$ 28 $\times$ 28, to the tensor. All of the noise is sampled from a normal distribution. We have empirically demonstrated that this method brings significant performance improvement despite its simplicity. Moreover, we found that adding channel-wise noise to the tokens in ViT-VQGAN, the tokenizer we used, effectively changes the colors of the decoded images, unlike adding Gaussian noise in entire dimensions. 
Example decoded images by ViT-VQGAN are presented in \cref{subsec:embnoise_vis_supp}.

\subsubsection{Stem-Adapter module}

As the tokens have a smaller size than images, they cannot be directly used for input of networks. We introduce a Stem-Adapter that converts the augmented tensor into ViT/16 to make minimal modifications on the network. Specifically, the Stem-Adapter module converts the 32 $\times$ 28 $\times$ 28 pre-processed tokens into 768 $\times$ 14 $\times$ 14, the same as the input of transformer blocks of ViT after the stem layer. 
We implement the Stem-Adapter module as a convolutional layer with a kernel size of 4 and a stride of 2. This allows the module to capture the spatial relationships of adjacent tokens and produce a tensor that can be used as input to ViT. The comparison among the different Stem-Adapter architectures is included in Section \cref{sec:ablation}.

\section{Experiments}
In this section, we conduct various experiments to demonstrate the effectiveness of token-based training. First, we compare \ours with four image compression methods on ImageNet-1k \cite{russakovsky2015imagenet}. Next, we explore the potential of \ours as a large-scale pre-training dataset by employing the ImageNet-21k dataset. We also provide ablation studies on the proposed token augmentation methods and Stem-Adapter module to determine the effectiveness of each proposed element. Lastly, we evaluate a continual learning scenario on the ImageNet-100 \cite{imagenet100} dataset to demonstrate the benefits of tokens in a limited memory environment.
The fine-grained classification results can be found in Appendix.

\subsection{ImageNet-1k classification}

ImageNet-1k classification performances are summarized in \cref{tab:main_results} and \cref{fig:teaser}.
Random sampling (\textcolor{YellowOrange}{yellow} in \cref{fig:teaser}) had the most significant negative impact on performance as storage capacity decreased. On the other hand, sampling by C-score~\cite{cscores} (\textcolor{Cyan}{blue}) also resulted in a noticeable performance drop, but it performed better than random sampling when storage capacity reduced to 10\% of the original. 
Although both sampling-based methods led to a considerable performance drop even with a small decrease in storage, JPEG-based compression methods (\textcolor{ForestGreen}{green}) maintained their performance until storage reached 50\% of the original.
When the quality was set above 50, the performance remained nearly the same as the original, even with 24.3\% of the original storage usage. However, when the quality was set to 1, the performance dropped dramatically to 67.8\%. Adjusting the resolution (\textcolor{Purple}{purple}) achieved better results than reducing the quality as storage became smaller while reducing the quality performed better than reducing the resolution with relatively large storage.
Despite the overall performance decline of image-based methods in low-storage environments, \ours achieved 74.0\% accuracy while using only 1\% of the original storage. Furthermore, by employing ImageNet-1k-5M \cite{yun2021re}, we were able to access more storage on tokens and achieve 78.6\% accuracy at 5\% of the ImageNet-1k storage size, where JPEG-based methods demonstrated performances lower than 75\%. These results highlight the effectiveness of \ours in improving performance in low-storage scenarios.

We also evaluate \ours model and the image-trained model on robustness datasets, such as adding Gaussian noise or Gaussian blur, ImageNet-R \cite{imagenet-r}, and adversarial attacks \cite{pgd, autoattack} in \cref{subsec:robustness_exp_supp}.
We observe that without strong pixel-level augmentations, \ours shows lower performance drops compared to the pixel-trained counterparts on corruptions and distribution shifts. \ours shows a significant gradient-based attack robustness compared to others.

\subsection{Storage-efficient token pre-training}
We extract tokens from ImageNet-21k dataset and pre-trained a ViT-B/16 model on the tokenized ImageNet-21k to determine the effectiveness of tokens as a large-scale pre-training. We then fine-tuned the pre-trained model with both tokenized ImageNet-1k and full-pixel ImageNet-1k, respectively (details are in \cref{subsec:hyperparam_exp_supp}).
Additionally, we extend our storage-efficient pre-training in three stages, namely, 21k token pre-training $\rightarrow$ 1k token pre-training $\rightarrow$ 1k image fine-tuning, following BeiT v2 \cite{peng2022beit2} (details are in \cref{subsec:storage_efficient_pt_exp_supp}).
The results are shown in \cref{tab:token_pretrain}.

The use of large-scale tokens for pre-training improved not only the performance of ImageNet-1k benchmarks using tokens but also the performance of full-pixel images. Pre-training with ImageNet-21k tokens led to a 2.5\% performance gain compared to using ImageNet-1k-5M tokens, using only 8GB more storage. Furthermore, our pre-training strategy improved full-pixel ImageNet-1k performance by 1.0\% using only 11.4\% more storage compared to the original full-pixel ImageNet-1k training.
It is only 27\% storage size compared to the sampling-based image pre-training strategy with a similar accuracy (410\% of IN-1k, showing 82.5\% accuracy) as shown in \cref{tab:pt_comparison}.

\begin{table}[t]
\small
\centering
\begin{tabular}{ccrrl}
\toprule

Pre-training & Fine-tuning  & \multicolumn{2}{c}{Storage}    & \multirow{2}{*}{Acc.} \\
IN-21k & IN-1k  & \multicolumn{1}{c}{Size} & \multicolumn{1}{c}{Ratio}    &  \\ 
\midrule
-         & Pixels & 140 GB & 100.0\% & 81.8$^\dagger$ \\
Tokens & Tokens & 16 GB & 11.1\% & 81.1   \\
Tokens & Pixels & 154 GB & 110.0\%& 82.6   \\
Tokens & Tokens $\rightarrow$ Pixels & 156 GB & 111.4\%& \textbf{82.8}   \\
\bottomrule
\end{tabular}
\vspace{.5em}
\caption{\small \textbf{Impact of storage-efficient pre-training (PT) and fine-tuning (FT).} We show the scenario of storage-efficient PT; we pre-train a model with a tokenized ImageNet-21k with more data points and fine-tune the model on the pixel or the token ImageNet-1k dataset. $^\dagger$ is from the original paper. ``Tokens $\rightarrow$ Pixels'' denotes three-staged FT, Token 21k PT, Token 1k PT and Pixels FT.
}
\label{tab:token_pretrain}
\end{table}

\begin{table}[t]
\small
\centering
\begin{tabular}{c|ccccc}
\toprule
\# PT images & $\times$1.35 & $\times$1.70 & $\times$2.05 & $\times$2.40 & $\times$3.10 \\
IN-1k FT Acc & 79.1 & 81.4 & 81.0 & 80.9 
 & 82.5 \\ \bottomrule 
\end{tabular}
\vspace{.5em}
\caption{\small {\bf Sampling-based pixel PT.} We show the IN-1k FT accuracies by different PTs by subsampling ImageNet-1k-5M \cite{yun2021re}. Pixel-based PT-FT strategy shows comparable accuracy to \ours when 410\% storage size is used (82.5 and 82.8, respectively).}
\label{tab:pt_comparison}
\end{table}

\subsection{Ablation study}
\label{sec:ablation}
We present an analysis of the proposed augmentation methods, Stem-Adapter architectures, and results on convolutional networks. \cref{tab:ablation_augmentation} reports the impact of the proposed augmentations for tokens. 
We found that employing Token-CutMix not only stabilized the overall training procedures but also resulted in the largest performance gain (8.1\%) compared to excluding it. 
The newly proposed methods for tokens, Embedding-Noise and Token-EDA, also showed performance improvements of 0.3\% and 1.4\%, respectively. Interestingly, these methods not only work effectively when used individually but also achieve higher performance when used in combination (74.0\%).

We also assessed the impact of the Stem-Adapter architecture on performance in \cref{tab:ablation:stem}. We compared two different Stem-Adapter architectures with our design choice. Note that, we used a smaller learning rate of 0.0005 for the linear Step-Adapter because of its unstable convergence using a larger learning rate and an input size of 14 $\times$ 14 to match the number of input patches with the convolutional Stem-Adapters. The results validate that our decision to use Conv $4\times4$ as Stem-Adapter for ViT models yields the highest performance among the considered candidates.

We also investigated the applicability of \ours to convolutional networks. The benchmark results on different architectures of ImageNet-1k are presented in \cref{tab:results_other_arch}. Note that token-based training only requires 1.4GB storage, which is merely 1\% of the storage required for pixel-based training. To match the size of features after the stem layer, we used a deconvolutional Stem-Adapter for ResNet \cite{resnet} models. Our findings indicate that \ours can also be used for storage-efficient training of convolutional models.

Finally, we show the impact of the tokenizer in \cref{subsec:more_tokenizers_exp_supp}. In summary, we observe that \ours works well for various tokenizers, \eg, ViT-VQGAN \cite{vitvqgan} and VQGAN \cite{vqgan} variants. We chose ViT-VQGAN considering the trade-off between the performance and the storage size, and it is solely trained on ImageNet-1k without external datasets.

\begin{table}[t]
\small
\centering
\begin{tabular}{@{}cccccc@{}}
\toprule
Token-CutMix & Token-EDA & Emb-Noise & Acc. (ViT-B)\\
\midrule
\nomark & \nomark & \nomark & 63.8 \\
\yesmark & \nomark & \nomark & 71.9 \\
\yesmark & \yesmark & \nomark & 72.2 \\
\yesmark & \nomark & \yesmark & 73.3 \\
\yesmark & \yesmark & \yesmark & \textbf{74.0} \\
\bottomrule
\end{tabular}
\vspace{.5em}
\caption{\small \textbf{Impact of the proposed augmentations.} ImageNet-1k validation accuracies for the combination of the proposed augmentations for tokens are shown.}
\label{tab:ablation_augmentation}
\end{table}

\begin{table}[t]
\small
\centering
\begin{tabular}{lccc}
\toprule
 & Linear & Conv $2\times2$ & Conv $4\times4$  \\
\midrule
Accuracy & 58.6 & 73.1 & \textbf{74.0} \\
\bottomrule
\end{tabular}
\vspace{.5em}
\caption{\small \textbf{Stem-Adapter architectures.} We compare three Stem-Adapter architectures for ViT-B/16 on ImageNet-1k. Note that stride of Convolution layers set to 2.}
\label{tab:ablation:stem}
\end{table}

\begin{table}[t]
\small
\centering
\begin{tabular}{ccc|cc}
\toprule
\multirow{2}{*}{Network} & \multicolumn{2}{c|}{Pixel-based training} & \multicolumn{2}{c}{Token-based training} \\
 & Acc. & Storage & Acc. & Storage \\
\midrule
ViT-S \cite{vit} & 79.9 & 140GB & 73.5 & 1.4GB \\
ResNet-50 \cite{resnet} & 76.1 & 140GB & 67.7 & 1.4GB \\
ResNet-18 & 69.7 & 140GB & 58.0 & 1.4GB \\
\bottomrule
\end{tabular}
\vspace{.5em}
\caption{\small \textbf{Comparisons on various architectures.} We additionally compare the performances of the pixel-training and token-training accuracies of three architectures, including ViT-S, ResNet-50, and ResNet-18, on the ImageNet-1k benchmark.}
\label{tab:results_other_arch}
\end{table}

\subsection{Continual learning}
\begin{figure}
    \centering
    \includegraphics[width=\linewidth]{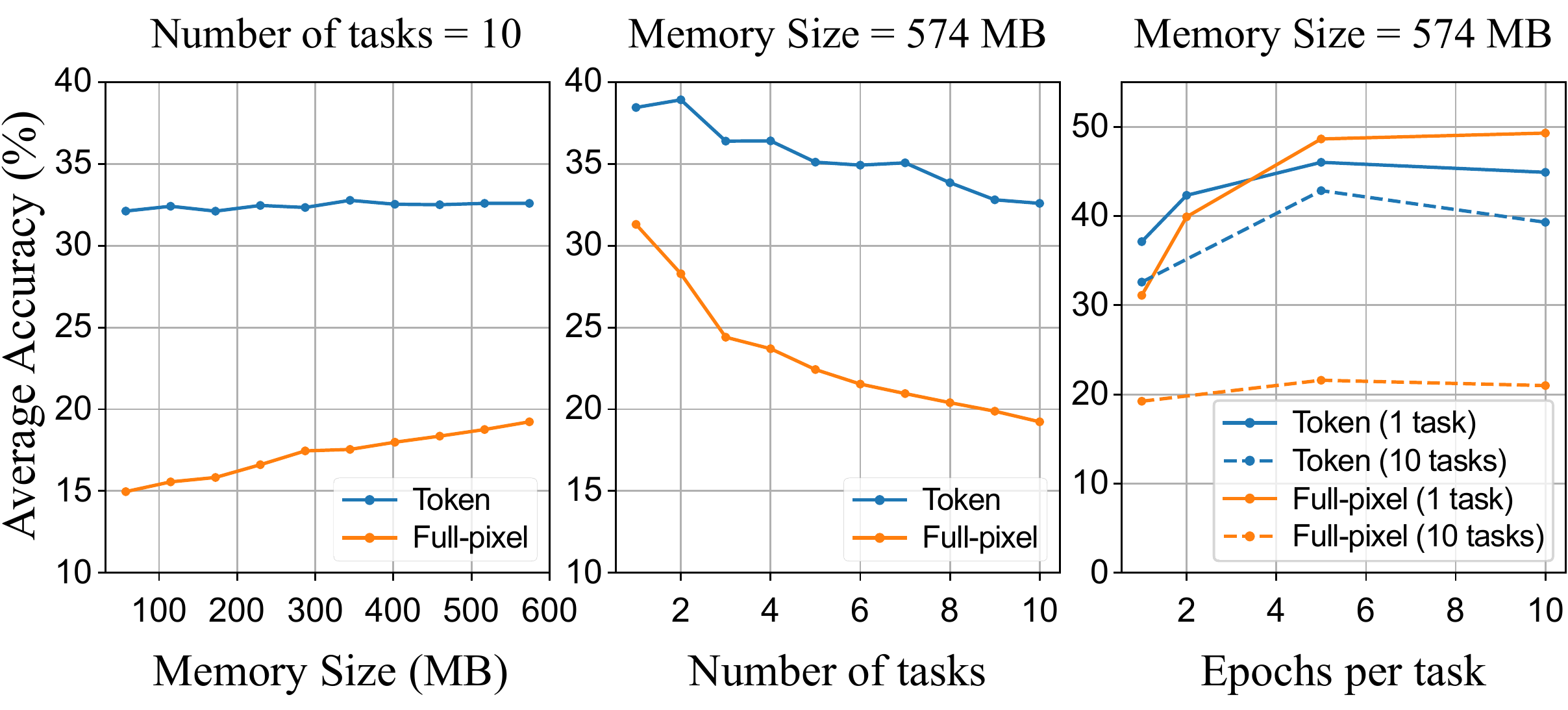}
    \caption{\small \textbf{Comparisons on the continual learning task.} We train two Experience Replay (ER) \cite{rolnick2019experience_replay_er} models on the ImageNet-100 \cite{imagenet100} dataset using the pixel dataset and the token dataset. (a) By varying the memory size while the number of tasks is fixed by 10. (b) By varying the number of tasks while fixing the memory size. (c) By increasing the epochs per task. Note that except (c), we set the epochs per task to 1 following the original setting \cite{rolnick2019experience_replay_er}.}
    \label{fig:continual_learning}
\end{figure}

To demonstrate the effectiveness of \ours in memory-limited settings, we compare \ours with full-pixel datasets in a continual learning scenario. Specifically, we employed the Reduced ResNet-18 architecture on the ImageNet-100 dataset~\cite{imagenet100} and evaluated the results following the Experience Relay~\cite{rolnick2019experience_replay_er}. 
We observed that when using the same memory size, \ours is significantly more memory-efficient than images, with a storage capacity of 147 times that of images. As a result, the total memory required to store the entire dataset in tokens was less than 500MB.
\cref{fig:continual_learning} illustrates the comparison results between using a token dataset and a full-pixel dataset in three different settings.

The left figure shows the performances of the token dataset and the full-pixel dataset by increasing memory size while fixing the number of tasks to ten. \ours outperforms the pixel dataset and shows a neglectable performance drop even when the memory size decreased, as it stored sufficient data even with memory sizes below 100MB.

The center figure presents the results of changing the number of tasks with a fixed memory size of 574MB ($\approx$ 1k images). In this case, both token and full-pixel datasets exhibited decreased performance as the number of tasks increased. However, the performance degradation of the token dataset was less severe than that of the full-pixel dataset.

Finally, with both memory size and the number of tasks fixed, we varied the number of times the dataset was viewed per task (the right figure). When there was only one task, the full-pixel dataset outperformed the token dataset as the epoch increased, consistent with other classification benchmark results. However, when there were ten tasks, the full-pixel dataset had lower performance than the token dataset, even with increased epochs due to insufficient stored data.

\subsection{Implementation details}
We used a pre-trained ViT-VQGAN Base-Base \cite{vitvqgan} model for extracting tokens from the images. Extracting tokens of entire ImageNet-21k dataset took 1.1 hours using 64 A100 GPUs with 2048 batch-size.
We conducted ImageNet-1k benchmark experiments using the ViT-B/16 model \cite{vit, deit} with an input size of 224 x 224. 
For token ImageNet-1k training, we replaced the patch embedding layer in ViT-B/16 model with the proposed Stem-Adapter module and added a global pooling layer before the final norm layer for tokens. 
We used a learning rate of 0.0015 with cosine scheduling and a weight decay of 0.1. The model was trained for 300 epochs with a batch size of 1024. We followed the training recipe proposed in DeiT~\cite{deit} for remaining settings except for the data augmentations.
We also followed the training recipe proposed in DeiT for the full-pixel ImageNet-1k training but made a few adjustments to handle the reduced datasets. We used a smaller learning rate of 0.0009 with a batch size of 1024 compared to the original value of 0.001, as we found that the original learning rate did not converge well on smaller datasets. Also, we increased the number of warm-up epochs and total training iterations when the number of data points decreased to ensure a fair comparison.
For large-scale token pre-training and token fine-tuning, we adopted simple augmentation strategies as suggested in DeiT-III \cite{touvron2022deit}; we excluded Token-EDA and replaced RRC with a simple random crop. 
Following the DeiT-III training recipe, we pre-trained the model with tokenized ImageNet-21k dataset for 270 epochs and then we fine-tuned the model for 100 epochs both of token and full-pixel dataset using learning rates of 0.00001 with 4096 batch-size and 0.0005 with 1024 batch-size, respectively.
We provide the more detailed hyper-parameter setting of our experiments in \cref{subsec:hyperparam_exp_supp}.

\section{Conclusion}

In this paper, we propose \oursfull by storing images into tokens. In practice, we store an image into 1kB as a 32$\times$32 token sequence and propose an efficient and fast encoding and decoding strategy for the token data type. We also propose token augmentations and Stem-Adaptor to train vision transformers with minimal modifications from the highly-optimized pixel-based training. Our experiments show that compared to the other storage-efficient training methods, \ours shows significantly large gaps; with the same amount of storage size, \ours shows the best performance among the comparison methods. Our method also shows benefits in other practical scenarios, such as storage-efficient large-scale pre-training and continual learning at scale.

\appendix
\numberwithin{equation}{section}
\numberwithin{figure}{section}
\numberwithin{table}{section}

\section*{Appendix}

In this additional document, we describe more details of \ours in \cref{sec:more_details_supp}, including the details of token encoding-decoding algorithms (\cref{subsec:pseudocode_supp}), the visualization of Emb-noise augmented tokens (\cref{subsec:embnoise_vis_supp}). We also include additional experimental results in \cref{sec:more_exp_supp}, including the hyperparameter details (\cref{subsec:hyperparam_exp_supp}), the full experimental results of Table 2 (\cref{subsec:full_exp_supp}), the additional results on storage-efficient pre-training (\cref{subsec:storage_efficient_pt_exp_supp}), exploring other tokenizers (\cref{subsec:more_tokenizers_exp_supp}), and robustness benchmarks (\cref{subsec:robustness_exp_supp}).

\section{More Details for \ours}
\label{sec:more_details_supp}

\subsection{Pseudo-code for Token Encoding-Decoding}
\label{subsec:pseudocode_supp}

\begin{algorithm}[b]
\caption{An algorithm for token encoding}\label{alg:encoding_supp}
\begin{algorithmic}[1]
\Require A sequence of tokens $T = [t_1, \ldots, t_N]$, the bits for the storage $M$
\State $L_{T} \gets [\phi]$
\Comment{Initialize an empty list for tokens}
\State $L_\text{idx} \gets [\phi]$
\Comment{Initialize an empty list for start indices}
\State $j \gets 0$
\While{$i \leq$ N}
\If{$t_i \geq 2^M$}
    \State $L_{T}$\texttt{.append} ($2^M$)
    \State $L_{T}$\texttt{.append} ($t_i - 2^M$)
    \State $j \gets j + 2$
\Comment{Assume $t_i < 2^{M+1}$ for simplicity}
\Else
    \State $L_{T}$\texttt{.append} ($t_i$)
    \State $j \gets j + 1$
\EndIf
\State $L_\text{idx}$\texttt{.append} ($j$)
\State $i \gets i + 1$
\EndWhile \\
\textbf{Return:} \texttt{Huffman encoding}~($L_{T}, L_\text{idx}$)
\end{algorithmic}
\end{algorithm}

\begin{algorithm}[t]
\caption{An algorithm for token decoding}\label{alg:decoding_supp}
\begin{algorithmic}[1]
\Require A compressed bytestring $L_{T}\prime$ from \cref{alg:encoding_supp}
\State $L_T = [t_0, \ldots, t_N], L_\text{idx} \gets$ \qquad \qquad \qquad \qquad \, \texttt{Huffman deencoding}~($L_{T}^\prime, L_\text{idx}^\prime$)
\State $T \gets [\phi]$
\State $i \gets 0$
\While{$L_\text{idx}$ is not empty}
\State $j \gets L_\text{idx}$\texttt{.pop} (0)
\State $k \gets i$
\While{$k \leq j$}
\If{$t_k \geq M$}
\State $T$\texttt{.append} ($t_k + t_{k+1}$)
\State $k \gets k + 2$
\Else
\State $T$\texttt{.append} ($t_k$)
\State $k \gets k + 1$
\EndIf
\EndWhile
\State $i \gets j$
\EndWhile \\
\textbf{Return:} $T$
\end{algorithmic}
\end{algorithm}

\cref{alg:encoding_supp} and \cref{alg:decoding_supp} describe the psuedo-codes for the proposed token encoding and decoding. Here, we assume $\max t_i < 2^{M+1}$ for the simplicity. For example, in our main experiments, each token belongs to 391 classes and we set $M=8$, hence, $\max t_i = 391 < 2^{8+1} = 512$. If the number of token classes is larger than $2^{M+1}$, then our algorithm can be naturally extended by repeating line 6-7 in \cref{alg:encoding_supp}. By this simple algorithm, we achieved a nearly optimal compression ratio (1.11 kB vs. 1.08 kB per image) where almost 0.63 smaller than the 16-bit encoding (2.0 kB per image). Note that, we use the native \texttt{gzip} library to perform Huffman encoding and decoding for simplicity.

\begin{figure}[ht]
    \centering
    \includegraphics[width=0.8\linewidth]{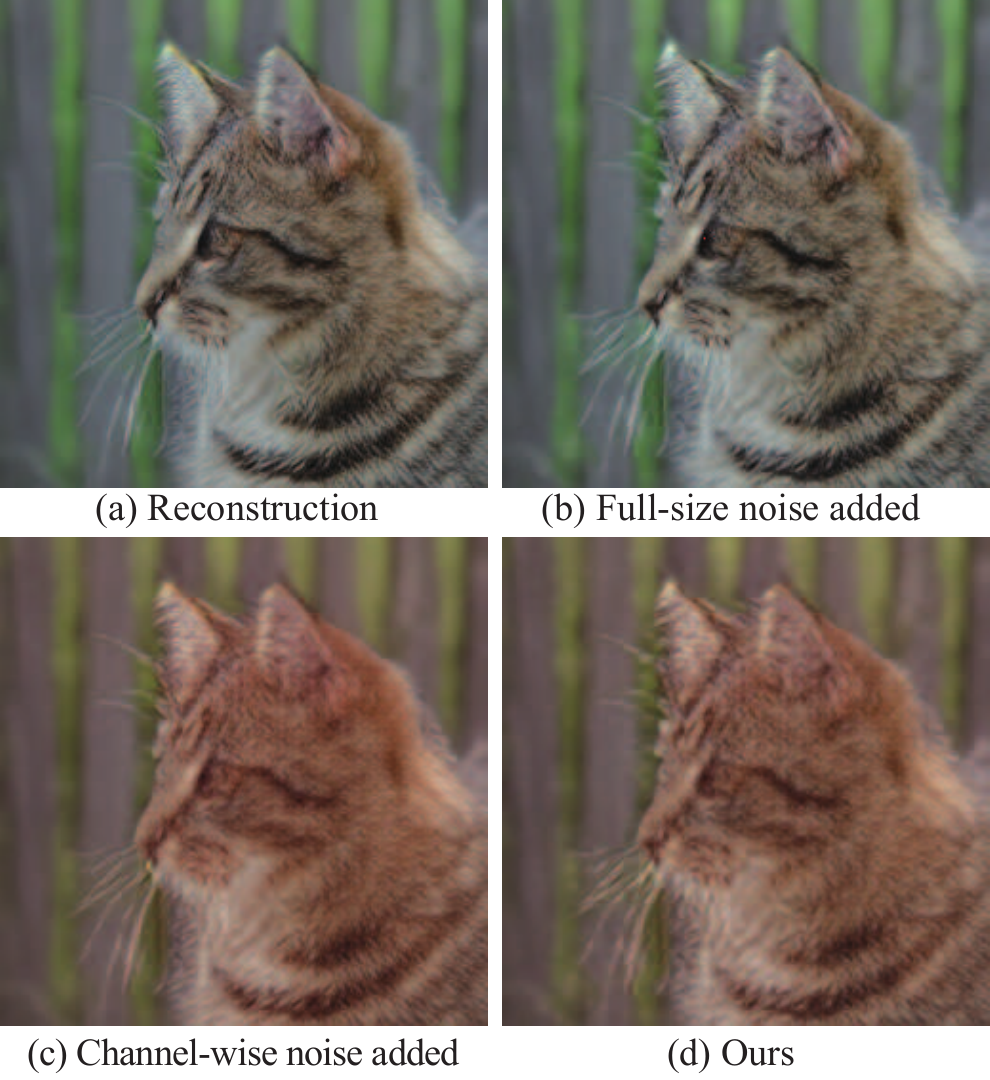}
    \caption{\small {\bf Emb-Noise visualization.} ``Reconstruction'' denotes the reconstructed image by the ViT-VQGAN decoder from the extracted tokens. ``Full-size noise'' is a random noise whose size is equivalent to the embedding vectors.}
    \label{fig:emb_noise_decode_overview_supp}
\end{figure}

\begin{figure*}[ht]
    \centering
    \includegraphics[width=\linewidth]{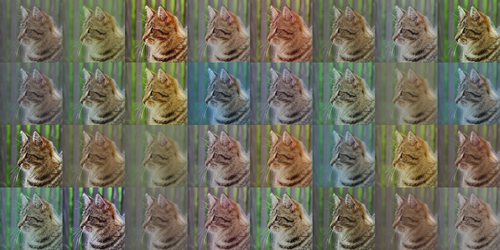}
    \caption{\small {\bf Channel-wise modification visualization.} We present ViT-VQGAN decoded images obtained by adding a constant to each of the 32 channels in codebook vectors.}
    \label{fig:emb_noise_decode_details_supp}
\end{figure*}

\subsection{ViT-VQGAN decoded images for Emb-Noise}
\label{subsec:embnoise_vis_supp}

\cref{fig:emb_noise_decode_overview_supp} and \cref{fig:emb_noise_decode_details_supp} show the visualization examples of the Emb-Noise augmented tokens and the tokens without augmentation. We use the ViT-VQGAN decoder for visualization. We observe that our Emb-Noise can make meaningful distortions on the decoded images.

\section{Additional Experimental Results}
\label{sec:more_exp_supp}

\subsection{Hyperparameter details}
\label{subsec:hyperparam_exp_supp}

\cref{tab:hyperparams_supp} shows the full list of hyperparameters used in our experiments. All hyperparameters are for the ViT-B backbones. In the table, Token IN-1k corresponds to \ours (ImageNet-1k) in Table 2, Token IN-21k PT corresponds to token pre-training in Table 3, and Token FT and Image FT correspond to token and image fine-tuning in Table 3, respectively. For other backbones and datasets, we only adjust the learning rate as the maximum learning rate showing a stable convergence (\eg, We use 0.15 for ResNet and ViT-S uses the same learning rate as ViT-B).

\begin{table*}[t]
\footnotesize
\centering
\begin{tabular}{lrrrrr}
\toprule
Methods             & DeiT IN-1k \cite{deit}      & Token IN-1k       & Token IN-21k PT & Token IN-1k FT & Image IN-1k FT \\ \midrule
Epochs              & 300             & 300              & 270     & 100           & 100           \\ \midrule
Batch size          & 1024            & 1024             & 2048    & 4096          & 512           \\
Optimizer           & AdamW           & AdamW            & AdamW   & AdamW         & AdamW         \\
Learning rate       & 0.0005 x $\frac{bs}{512}$ & 0.00075 x $\frac{bs}{512}$ & 0.0015  & 0.00001       & 0.0005        \\
Learning rate decay & cosine          & cosine           & cosine  & \bnomark      & cosine        \\
Weight decay        & 0.05            & 0.1              & 0.02    & 0.1           & 0.05          \\
Warmup epochs       & 5               & 5                & 5       & 5             & 5             \\
Label smoothing     & 0.1             & 0.1              & 0.1     & 0.1           & 0.1           \\
Dropout             & \bnomark        & \bnomark         & \bnomark  & \bnomark  & \bnomark        \\
Stoch. Depth        & 0.1             & 0.1              & 0.1     & 0.15          & 0.1           \\
Gradient Clip       & \bnomark        & \bnomark         & \bnomark  & \bnomark  & \bnomark        \\ \midrule
Cutmix prob.        & 1               & 1                & 1       & 1             & 1             \\
Mixup prob.         & 0.8             & 0                & 0       & 0             & 0.8           \\
RandAug             & 9 / 0.5         & -                & -       & -             & 9 / 0.5       \\
Repeated Aug        & \byesmark       & -                & -       & -             & \bnomark      \\
Erasing prob.       & 0.25            & -                & -       & -             & 0             \\
EDA prob.           & -               & 0.25 (RS) / 0.25 (SR)             & 0       & 0             & -             \\
Emb-Noise prob.     & -               & 0.5              & 0.5     & 0.5           & -             \\
\bottomrule
\end{tabular}
\vspace{.5em}
\caption{\small {\bf Hyperparamters for \ours and DeiT-B.} All hyperparameters are for the ViT-B backbone. DeiT IN-1k is the same as the original DeiT paper (baseline).}
\label{tab:hyperparams_supp}
\end{table*}

\subsection{The full experimental results}
\label{subsec:full_exp_supp}

We report the full experimental results in \cref{tab:main_results_supp}. Details are the same as Table 2. 

\begin{table}[ht]
\small
\begin{tabular}{lcccc}
\toprule
Pre-trained on              & Flowers & Cars & iNat18 & iNat19 \\
\midrule
Pixel (IN-1k)               & 98.0       & 91.8         & 73.0   & 77.7   \\
Token (IN-1k)               & 93.5       & 79.7         & 43.1   & 50.1   \\
Token (IN-21k, IN-1k)       & 98.7       & 84.5         & 50.1   & 58.3  \\
\bottomrule
\label{tab:otherdatasets}
\end{tabular}
\caption{\small {\bf Other datasets.} We report the top-1 accuracies on diverse fine-grained datasets achieved by \ours. We tested \ours on the Flowers \cite{flowers102}, StanfordCars \cite{stanfordcars}, iNaturalist (iNat)-18 \cite{inaturalist18} and iNat-19 \cite{inaturalist19} datasets.}
\end{table}

\begin{table*}[ht]
\small
\centering
\begin{tabular}{lllllll}
\toprule
Tokenizer & Training dataset & Quantiztation & Voca size (\# of valid voca) & PS & FID & ViT-S (SeiT) Acc \\ \midrule
VQGAN & ImageNet & Vector quantization & 1024 (454) & 16 & 7.94 & 75.3 \\
VQGAN & ImageNet & Vector quantization & 16384 (971) & 16 & 4.98 & 76.9 \\
VQGAN & OpenImages & Gumbel quantization & 8192 (2886) & 8 & 1.49 & 79.1 \\
VQGAN & OpenImages & Vector quantization & 256 (256) & 8 & 1.49 & \textbf{81.8} \\
ViT-VQGAN & ImageNet & Vector quantization & 8192 (391) & 8 & 1.28 & 77.3 \\
\bottomrule
\end{tabular}
\vspace{.5em}
\caption{\small {\bf Exploring other tokenizers.} Various ViT-S (\ours) results on the ImageNet-100 benchmark are shown. We compare various VQGAN tokenizers with ViT-VQGAN by varying the quantization methods (Gumbel softmax vs. vector quantization) the vocabulary size, the valid vocabulary size (the number of classes actually used for the ImageNet-1k training dataset), and the patch size (PS).}
\label{tab:other_tokenizers_supp}
\end{table*}

\begin{table*}[ht]
\small
    \centering
    \begin{tabular}{ccccccc}
    \toprule
    Model & Data format & Clean & Gauss. Noise & Gauss. Blur & ImageNet-R & Sketch \\ \midrule
    ViT-S (DeiT) & Pixels & 79.9  & 75.1 {\footnotesize (6.0\%)} & 73.4 {\footnotesize (8.1\%)} & 28.8 {\footnotesize (63.9\%)} & 29.9 {\footnotesize (62.6\%)} \\ \midrule
    ViT-S (Weak Aug) & Pixels & 78.0 & 64.7 {\footnotesize \textbf{(17.1\%)}} & 66.8 {\footnotesize (14.4\%)} & 20.8 {\footnotesize (73.4\%)} & 18.1 {\footnotesize (76.8\%)} \\
    ViT-S (SeiT, ours) & Tokens & 74.0  & 60.8 {\footnotesize (17.3\%)} & 65.3 {\footnotesize \textbf{(11.2\%)}} & 26.0 {\footnotesize \textbf{(64.6\%)}} & 23.0 {\footnotesize \textbf{(68.7\%)}} \\
    \bottomrule
    \end{tabular}
    \vspace{.5em}
    \caption{\small \textbf{Robustness evaluation.} We show the clean and robust accuracies against corruptions and domain shifts of each model trained on ImageNet-1k. The performance drops are put in parentheses (lower is better) for robust accuracies.}
    \label{tab:robustness}
\end{table*}

\subsection{Other datasets}
\label{subsec:other_datasets}

The performances of \ours on various datasets are reported in Table \ref{tab:otherdatasets}.
We tokenize the datasets and then fine-tune a model (ViT-B) with the tokenized dataset, using token-trained model weights. The token-trained models weights, Token (IN-1k) and Token (IN-21k, IN-1k) achieve top-1 accuracies of 74.0\%, 81.1\% on ImageNet-1k, respectively.
The pixel counterpart is fine-tuned on pixel target datasets from the pixel pre-trained model; Pixel (IN-1k), showing 81.8\% top-1 accuracy on IN-1k.
We followed the pixel-training recipes of DeiT \cite{deit}. Although we do not modify the training recipe for tokens, the results verify the possibility of SeiT on those datasets.

\begin{figure}
    \centering
    \includegraphics[width=.8\linewidth]{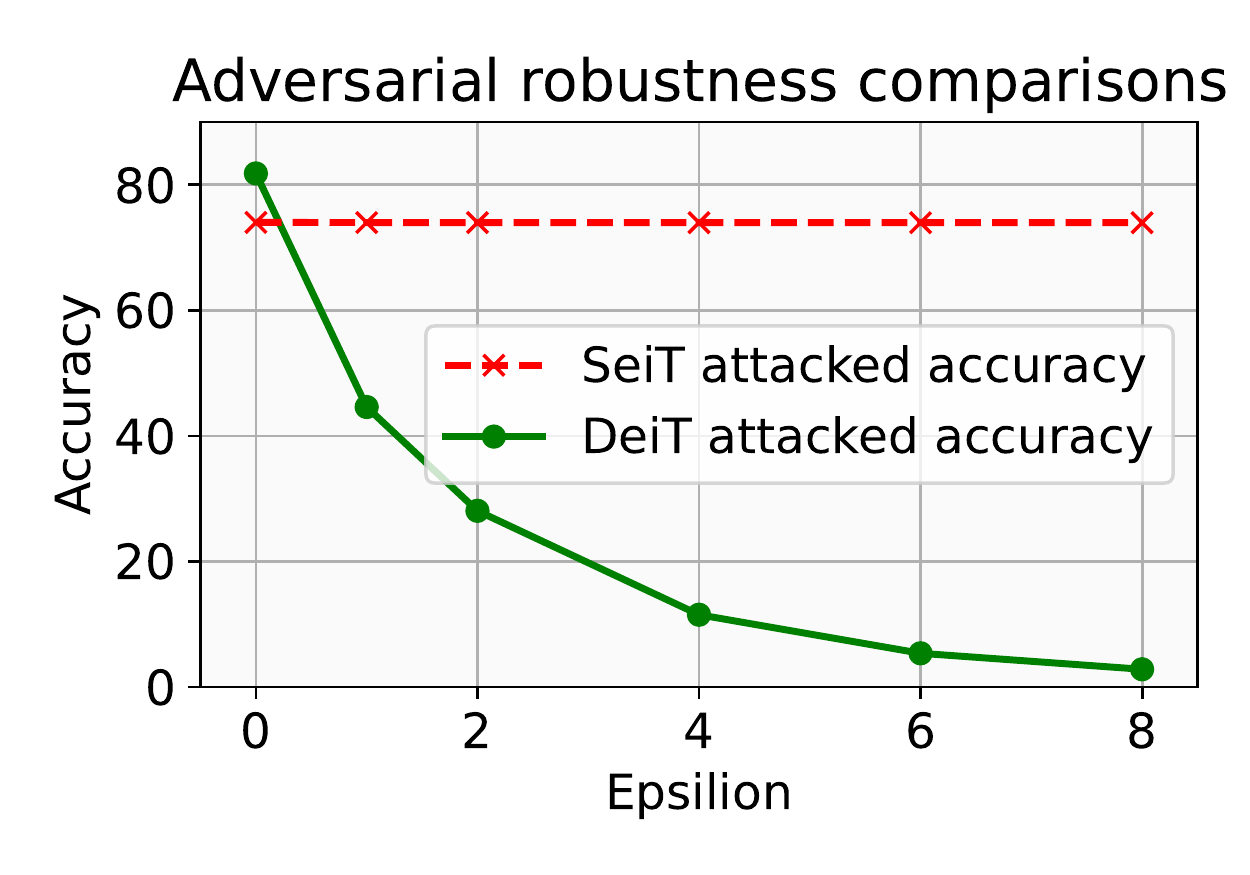}
    \caption{\small {\bf Adversarial robustness of DeiT and SeiT by varying $\varepsilon$.} $\varepsilon=0$ denotes the clean accuracy.}
    \label{fig:advattack_supp}
\end{figure}

\subsection{Three-stage storage-efficient pre-training}
\label{subsec:storage_efficient_pt_exp_supp}

Following BeiT v2 \cite{peng2022beit2}, we extend our storage-efficient pre-training in three stages, namely, 21k token pre-training $\rightarrow$ 1k token pre-training $\rightarrow$ 1k image fine-tuning. For simplicity, we directly fine-tune the ``21k token pre-trained and 1k token fine-tuned model'' (\ie, 81.1\% model in Table 3) on the image pixels with the same optimization hyperparameter of the image fine-tuned model. As a result, we have 82.8\% top-1 accuracy, slightly better than the original two-staged training strategy (+0.2\% than 82.6\%).

\subsection{Exploring other tokenizers}
\label{subsec:more_tokenizers_exp_supp}

In this subsection, we explore other tokenizers rather than ViT-VQGAN \cite{vitvqgan}, \eg, VQGAN \cite{vqgan}. We employ four VQGAN models from the official repository\footnote{\url{https://github.com/CompVis/taming-transformers}}, ImageNet-trained VQGAN with patch size 16 and vocabulary size 1024, ImageNet-trained VQGAN with patch size 16 and vocabulary size 16384, OpenImages \cite{openimages}-trained VQGAN with patch size 8 and vocabulary size 256, and OpenImages \cite{openimages}-trained VQGAN with patch size 8 and vocabulary size 8192. Here, the last VQGAN model is trained with the Gumbel softmax \cite{gumbel1,gumbel2} quantization, instead of the original vector quantization by VQ-VAE \cite{vqvae}. Here, we slightly change our Stem-Adopter from 4$\times$4 Conv with stride 2 to 2$\times$2 Conv with stride 1 for tokenizers with patch size 16.

In \cref{tab:other_tokenizers_supp}, we report the ViT-S (\ours) top-1 accuracy on the ImageNet-100 benchmark by varying the choice of tokenizers. We also report the reported ImageNet-1k validation FID score of each tokenizer. In the table, we observe that the top-1 accuracy of \ours follows the generation quality (FID) if we use the same quantization method (\eg, vector quantization). The ViT-VQGAN shows the best FID (1.28) as well as the best ImageNet performance with \ours (77.3). While the Gumbel quantized VQGAN achieves the best performance, in practice, we use ViT-VQGAN due to two reasons. First, the storage efficiency: 2886 valid codes need 1.5 times more storage than 391 valid codes. Second, Although the OpenImages \cite{openimages}-trained VQGAN shows better quality, it needs to be trained on a large-scale external dataset. We did not use the OpenImages-trained VQGAN for a fair comparison with other ImageNet-1k-only training methods.

\subsection{Robustness benchmarks}
\label{subsec:robustness_exp_supp}

We compare ViT-S models trained on ImageNet-1k with different training strategies using robustness benchmarks. We employ three scenarios: (1) noise and blur scenario (2) domain shift scenario (3) adversarial attack scenario. For the first scenario, we add Gaussian noise and Gaussian blur to the validation images. We use ImageNet-R \cite{imagenet-r} and Sketch-ImageNet \cite{sketchimagenet} for testing the robustness against domain shifts. Finally, we use a weak version of AutoAttak \cite{autoattack} for measuring adversarial robustness.

As the original DeiT is trained on strong augmentation, such as RandAugment or 3-Augment, we also compare our method with ``weak augmented'' ViT-S, where it only employs resized random crop (RRC) and CutMix \cite{yun2019cutmix}. Our assumption is that because the pixel-trained models are sensitive to imperceptible details, they will be less robust than our approach in noise or adversarial attack scenarios. However, on the other hand, because our method relies on the encoding power of the pre-trained tokenizer, if the employed tokenizer is not a robust feature extractor, our method could be more vulnerable than pixel-trained counterparts.

\cref{tab:robustness} shows the results of the first and the second scenarios. Here, we observe two important findings. First, when we use the same augmentations with the same strength (ViT-S Weak Aug vs. ViT-S SeiT), \ours shows smaller performance drops on both noise scenarios and domain shift scenarios. On the other hand, when we use strong pixel-level augmentations, the pixel-trained counterpart outperforms our approach. It implies that the key to the input pixel robustness depends on the pixel-level augmentations with severe distortions as observed by previous studies \cite{chun2019icmlw, taori2020measuring}. However, because our method uses only tokens, not pixels directly, investigating how to explore pixel-level distortion augmentations on the token level will be an open question and an interesting future research direction.

We also compare the adversarial robustness of DeiT-S and SeiT-S. We employ the APGD (a step size-free version of PGD attack \cite{pgd}) with cross-entropy loss and DLR loss, following AutoAttack \cite{autoattack}. Because \ours employs discrete non-differentiable representations in the computational graph, we employ the straight-through estimator (STE) \cite{ste} to estimate the non-differentiable gradients, following Athalye \etal \cite{athalye2018obfuscated}. We also evaluate the non-quantized version of the quantizer (\ie, omitting the vector quantization process, but using the extracted feature by the encoder directly to the ViT input), but we empirically observe that attacking the non-quantized version cannot drop the performance at all. Instead, we use the STE, also used during the training as well as the previous extensive robustness study \cite{athalye2018obfuscated}. We compare the attacked accuracies of DeiT and SeiT by varying $\varepsilon$ (a control parameter for the attack intensity) from 0 to 8 in \cref{fig:advattack_supp}. We observe that \ours shows almost neglectable performance drops even under the strongest attack (showing 73.95 for $\varepsilon = 8$ where 73.98 for $\varepsilon = 0$), where DeiT shows 2.8\% top-1 accuracy.

However, we should be careful to interpret \cref{fig:advattack_supp}; it could be due to a strong obfuscated gradient effect \cite{athalye2018obfuscated} that cannot be detected by a naive straight-through estimator. Moreover, our method could be vulnerable to the codebook attack by changing the token indices directly, not by perturbing the pixels. However, as an efficient and natural adversarial attack on discrete domains is still an open problem \cite{zhang2020adversarial} (\eg, altering indices as imperceptible to humans but sensitive to machines --- only a small index change can make a huge semantic gap, such as replacing ``huge'' in the previous sentence to ``neglectable''), we leave the investigation of advanced adversarial attack methods for \ours beyond straight-through estimator as future work.

\begin{table*}[ht!]
\centering
\resizebox{\linewidth}{!}{
\begin{tabular}{ccrrrr}
\toprule
\multicolumn{2}{c}{Method}                        & Storage size & \# of images & Top1 Acc. \\
\midrule
\multicolumn{1}{c}{Full-pixels}           & 100\% & 140 GB                & 1.28 M         & 81.8      \\
\midrule
\multirow{8}{*}{Uniform random sampling}   & 20\% & 27.2 GB              & 0.26 M        & 59.8      \\
                                           & 30\% & 41.0 GB              & 0.38 M        & 69.3      \\
                                           & 40\% & 54.6 GB              & 0.51 M        & 74.0        \\
                                           & 50\% & 68.4 GB              & 0.64 M        & 76.0        \\
                                           & 60\% & 82.0 GB              & 0.77 M        & 77.8      \\
                                           & 70\% & 95.7 GB              & 0.90 M        & 78.2      \\
                                           & 80\% & 109.3 GB             & 1.02 M        & 79.4      \\
                                           & 90\% & 123.1 GB             & 1.15 M        & 81.1      \\
                                           \midrule
\multirow{8}{*}{C-score based sampling}    & 20\% & 26.3 GB              & 0.26 M        & 65.1      \\
                                           & 30\% & 39.8 GB              & 0.38 M        & 69.4      \\
                                           & 40\% & 53.3 GB              & 0.51 M        & 73.3      \\
                                           & 50\% & 66.9 GB              & 0.64 M        & 76.9      \\
                                           & 60\% & 80.6 GB              & 0.77 M        & 77.5      \\
                                           & 70\% & 94.3 GB              & 0.90 M        & 79.2      \\
                                           & 80\% & 108.1 GB             & 1.02 M        & 80.4      \\
                                           & 90\% & 121.8 GB             & 1.15 M        & 80.9      \\
                                           \midrule
\multirow{9}{*}{Adjusting image reolution} & 10\% & 5.3 GB                & 1.28 M        & 63.3      \\
                                           & 20\% & 9.6 GB                & 1.28 M        & 75.2      \\
                                           & 30\% & 16 GB                 & 1.28 M        & 78.6      \\
                                           & 40\% & 24 GB                 & 1.28 M        & 79.4      \\
                                           & 50\% & 34 GB                 & 1.28 M        & 80.9      \\
                                           & 60\% & 46 GB                 & 1.28 M        & 80.8      \\
                                           & 70\% & 60 GB                 & 1.28 M        & 81.6      \\
                                           & 80\% & 75 GB                 & 1.28 M        & 81.6      \\
                                           & 90\% & 93 GB                 & 1.28 M        & 80.8      \\
                                           \midrule
\multirow{9}{*}{Adjusting JPEG quality factor} & 1    & 9.3 GB                & 1.28 M        & 67.8      \\
                                           & 5    & 11 GB                 & 1.28 M        & 74.6      \\
                                           & 10   & 14 GB                 & 1.28 M        & 78.1      \\
                                           & 25   & 23 GB                 & 1.28 M        & 80.7      \\
                                           & 50   & 34 GB                 & 1.28 M        & 81.1      \\
                                           & 75   & 50 GB                 & 1.28 M        & 81.5      \\
                                           & 85   & 66 GB                 & 1.28 M        & 81.3      \\
                                           & 90   & 79 GB                 & 1.28 M        & 80.9      \\
                                           & 95   & 113 GB                & 1.28 M        & 81.6      \\
                                           \midrule
\multicolumn{2}{c}{SeiT (ImageNet-1k, ours)}            & 1.36 GB                & 1.28 M         & 74.0        \\
\multicolumn{2}{c}{SeiT (ImageNet-1k-5M, ours)}         & 7.49 GB                & 5.83 M           & 78.6      \\
\multicolumn{2}{c}{SeiT (ImageNet-1k, OpenImages-VQGAN)}            & 1.36 GB                & 1.28 M         & 78.4        \\
\multicolumn{2}{c}{SeiT (ImageNet-1k-5M, OpenImages-VQGAN)}         & 7.49 GB                & 5.83 M           & 78.7      \\

\midrule
\multicolumn{2}{c}{SeiT (IN-21k tokens $\rightarrow$ 1k tokens, ours)}            & 16 GB                & 12.4 M         & 81.1        \\
\multicolumn{2}{c}{SeiT (IN-21k tokens $\rightarrow$ 1k tokens, OpenImages-VQGAN)}            & 16 GB                & 12.4 M         & 82.3        \\
\multicolumn{2}{c}{SeiT (IN-21k tokens $\rightarrow$ 1k pixels, ours)}            & 154 GB                & 12.4 M         & 82.6        \\
\multicolumn{2}{c}{SeiT (IN-21k tokens $\rightarrow$ 1k tokens $\rightarrow$ 1k pixels, ours)}            & 156 GB                & 12.4 M         & 82.8        \\
\bottomrule
\end{tabular}
}
\vspace{.5em}
\caption{\small \textbf{The full main results.} The full results of \cref{fig:teaser}, \cref{tab:main_results} and \cref{tab:token_pretrain}. The results in the rows denoted to OpenImages-VQGAN are obtained by utilizing OpenImages-trained VQGAN with patch size 8 and vocabulary size 256 as the tokenizer.}
\label{tab:main_results_supp}
\end{table*}

\clearpage

{
\bibliographystyle{ieee_fullname}
\bibliography{reference}
}

\end{document}